\documentclass[lettersize,journal]{IEEEtran}
\usepackage{amsmath,amsfonts}
\usepackage{amssymb,mathtools}
\usepackage{bm}
\usepackage{algorithmic}
\usepackage{algorithm}
\usepackage{array}
\usepackage[caption=false,font=normalsize,labelfont=sf,textfont=sf]{subfig}
\usepackage{textcomp}
\usepackage{stfloats}
\usepackage{url}
\usepackage{verbatim}
\usepackage{graphicx}
\usepackage{cite}
\usepackage{booktabs}
\usepackage{multirow}
\usepackage{xcolor}
\usepackage[hidelinks]{hyperref}
\usepackage{enumitem}
\usepackage{tikz}
\usetikzlibrary{arrows.meta,positioning,calc,fit,backgrounds}
\graphicspath{{figures/}}

\newcommand{\Real}{\mathbb{R}}

\begin{document}

\title{Spacetime Optimal-Transport Attention for\\
Visuo-Haptic Imitation Learning of Contact-Rich Manipulation}

\author{Yue~Feng,~\IEEEmembership{Member,~IEEE,}
        Weicheng~Huang,
        and~I-Ming~Chen,~\IEEEmembership{Fellow,~IEEE}
\thanks{Yue Feng and I-Ming Chen are with the Robotics Research Centre, School of Mechanical and Aerospace Engineering, Nanyang Technological University, Singapore 639798. Weicheng Huang is with Wings Robotics. E-mail: yue011@e.ntu.edu.sg, info@wingsrobotics.com, michen@ntu.edu.sg.}
\thanks{This manuscript is a preprint deposited on arXiv. \copyright~2026 The Authors.}}

\maketitle

\bstctlcite{IEEEtranBSTCTL}

\begin{abstract}
Contact-rich manipulation tasks such as tight-clearance insertion,
connector mating, polishing, and surface-conforming wiping remain
difficult for data-driven controllers because they couple discontinuous
contact dynamics, partial observability, and strict safety constraints.
No single sensing modality suffices: vision supplies global context
before contact, force/torque (F/T) feedback governs interaction after
contact, and proprioceptive pose provides a consistent kinematic
backbone. Most prior imitation-learning policies for contact-rich tasks
operate on uni- or bi-modal signals, and the few that fuse three
modalities typically adopt off-the-shelf attention modules with no
explicit prior on how attention mass should be distributed across
task-relevant regions. We present \textbf{Spacetime Optimal-Transport
Attention (SO-TA)}, a tri-modal fusion backbone that replaces
softmax-normalized patch attention by an entropy-regularized Optimal
Transport (OT) alignment between force--pose-derived sub-queries and
visual patches. Explicit marginal constraints act as a structured
inductive bias for contact-rich tasks, encouraging conditioning-aware
spatial selection that is stable across illumination, distractors, and
partial occlusion. SO-TA is paired with a diffusion-based sequence
policy mapping observation windows to pose-action chunks. We evaluate
SO-TA on three real-robot tasks: tight peg-in-hole assembly, BCM
wiring-connector insertion, and curved-surface mark erasing. With
$\sim\!200$ rollouts per condition, SO-TA reaches $100\%$ success on
tight peg-in-hole versus $93\%$ for cross-attention at matched
capacity, and retains $82.5\%$ success under illumination, distractor,
and partial-occlusion perturbations where a concatenation baseline
drops to $43.5\%$. OT-derived patch heatmaps and leave-one-out
modality-influence ratios provide interpretable, phase-dependent
diagnostics.
\end{abstract}


\section{Introduction}
\IEEEPARstart{C}{ontact-rich} manipulation tasks pervade industrial and
service-robot deployments. Representative examples include
tight-clearance insertion, connector mating, deburring, polishing,
surface-conforming wiping, and
sealing~\cite{suomalainen2022contactsurvey,tsuji2025contactrichsurvey}.
These tasks share three properties that make data-driven control hard.
First, the dynamics transition abruptly between free-space motion and
constrained contact, so a controller must regulate interaction force the
moment contact begins; deficiencies in sensing or control can result in
force spikes, surface damage, and loss of
repeatability~\cite{suomalainen2022contactsurvey}. Second, task-relevant
visual evidence is often a small region of the camera frame whose
location shifts rapidly across task phases. Third, the operating envelope
is safety-constrained, so a deployed policy must behave predictably under
modest distribution shift.

These properties motivate tri-modal sensing that fuses vision, F/T, and
proprioceptive pose. Vision provides global scene context and coarse
geometric cues before contact. F/T feedback becomes essential after
contact to regulate interaction and prevent force
spikes~\cite{weinberg2024inhand_survey,xie2025forceful_fm}. Pose signals
provide a consistent kinematic backbone for both execution and phase
progression. Recent imitation-learning (IL) systems based on diffusion- or
transformer-based sequence
policies~\cite{Chi-RSS-23,Zhao-RSS-23,shafiullah2022behavior} have
demonstrated that complex behaviors can be learned from a modest number
of demonstrations. Empirical progress on tri-modal IL, however, is held
back on two fronts. Open multimodal datasets remain dominated by vision
and proprioception, and synchronized corpora with calibrated F/T are
scarce~\cite{vuong2023open,mccarthy2025towards,fang2023rh20t}. More
fundamentally, the fusion module that builds the conditioning signal for
these policies has received comparatively little attention in the
contact-rich setting. Most existing systems either concatenate
per-modality features or adopt standard cross-attention off-the-shelf
from other
domains~\cite{Baltrusaitis2019MMSurvey,Du2022MMTsurvey,tsuji2025contactrichsurvey,urain2024dgm_multimodal}.
Concatenation discards spatial structure, while softmax-normalized
cross-attention places no explicit constraint on the mass distributed
across patches and is therefore sensitive to spurious saliency from
clutter, occlusion, and illumination changes. Recent surveys on IL for
contact-rich tasks and on forceful foundation models converge on the same
diagnosis: progress requires moving beyond plug-and-play attention toward
principled fusion that is phase-aware, modality-selective, and aligned
with the physics of
contact~\cite{tsuji2025contactrichsurvey,xie2025forceful_fm}.

This paper introduces \textbf{Spacetime Optimal-Transport Attention
(SO-TA)}, a tri-modal fusion backbone designed around the regularities of
contact-rich manipulation. Concretely:
\begin{itemize}[leftmargin=*]
\item We formulate force--pose-conditioned spatial aggregation as an
entropy-regularized Optimal Transport alignment between
force--pose-derived sub-queries and visual patches. Explicit marginal
constraints replace the softmax normalization of standard cross-attention
and act as a structured inductive bias for selecting task-relevant
evidence.
\item We combine SO-TA with a per-frame modal Transformer and a temporal
Transformer to produce a fused encoding compatible with diffusion-policy
backbones, and we expose two interpretability signals: OT-derived patch
heatmaps and leave-one-out modal influence ratios.
\item We evaluate SO-TA against concatenation and cross-attention baselines
on three real-robot tasks. SO-TA reaches $100\%$ success on tight
peg-in-hole, retains $82.5\%$ success under visual perturbation when
concatenation drops to $43.5\%$, and produces phase-consistent
interpretability maps on both insertion and erasing.
\end{itemize}
The rest of the paper reviews related work
(Sec.~\ref{sec:related}), states preliminaries
(Sec.~\ref{sec:prelim}), formulates the tri-modal IL setup
(Sec.~\ref{sec:problem}), details SO-TA (Sec.~\ref{sec:sota}),
describes the baselines (Sec.~\ref{sec:baselines}), and reports
experiments (Sec.~\ref{sec:experiments}).

\section{Related Work}
\label{sec:related}
\paragraph*{Imitation learning and diffusion policies}
Imitation learning (IL) avoids reward design by fitting a policy to
expert
demonstrations~\cite{argall2009survey,osa2018survey,ravichandar2020recent}.
Recent sequence
policies~\cite{Chi-RSS-23,Zhao-RSS-23,shafiullah2022behavior} predict
future action chunks from short observation windows, improving temporal
consistency over single-step behavior cloning. Diffusion
Policy~\cite{Chi-RSS-23} in particular has emerged as a competitive
backbone because conditional denoising naturally handles multimodal action
distributions. Surveys of IL for contact-rich manipulation and of deep
generative models for multimodal
demonstrations~\cite{tsuji2025contactrichsurvey,urain2024dgm_multimodal}
converge on a common observation: the action-producing backbone is now
relatively mature, while the upstream fusion network that builds its
conditioning signal remains the bottleneck under contact. We adopt
Diffusion Policy as the action-producing module and focus on this
upstream fusion step.

\paragraph*{Attention-based visuo-haptic fusion}
Multimodal Transformers and attention
layers~\cite{Vaswani2017,Tsai2019MMLTransformer,Baltrusaitis2019MMSurvey,Du2022MMTsurvey}
are the default tool for fusing heterogeneous modalities, and several
systems have adopted them off-the-shelf for contact-rich IL. SoftGrasp
uses cross-modal self-attention for dexterous
grasping~\cite{LiSoftGrasp2024}; diffusion-policy approaches for
compliant manipulation insert self- or cross-attention between vision and
F/T tokens~\cite{AburubDiffusionCompliant2024,KangPryingDiffusion2024};
broader sensory-fusion architectures combine vision, audio, and touch
through attention~\cite{LiSeeHearFeel2023}; and Visuo-Tactile Transformers
extend the formulation to tactile streams~\cite{ChenVTT2023}. Earlier
work also demonstrated the value of vision--touch self-supervised
representations for contact-rich
tasks~\cite{lee2019makingsense,kalashnikov2018qtopt}. These designs
improve over concatenation but inherit the softmax normalization of
generic attention, which leaves the spatial selection step inside the
visual stream unconstrained and therefore vulnerable to spurious saliency
from clutter, occlusion, and illumination
changes~\cite{tsuji2025contactrichsurvey,xie2025forceful_fm}.

\paragraph*{Bimodal and force-centric methods}
A second line of work emphasizes force supervision and compliant control
rather than fusion design. Hybrid trajectory--force IL stabilizes assembly
with force supervision but omits vision~\cite{WangHybridTrajForce2023};
learned force control for paper wrapping concatenates transformed streams
without cross-modal alignment~\cite{HanaiPaperWrapping2023}; MOMA-Force
adds force to a visual policy for mobile manipulation but uses it
primarily for downstream compliant execution rather than as a token in a
tri-modal stack~\cite{yang2024momaforce}; ForceMimic introduces a
force--motion capture pipeline and reports strong insertion performance,
while its fusion step largely concatenates the
streams~\cite{LiuForceMimic2024}. These results confirm that force
supervision is valuable, but they leave the question of how to fuse three
modalities at the representation level largely open.

\paragraph*{Architecture- and stage-guided priors}
A third line of work encodes task structure directly into the fusion
model. Hierarchical and curriculum methods tailored to peg-in-hole inject
staged contact difficulty and perception--control
decoupling~\cite{JinHierFusion2024,JinVisionForceCurriculum2024}, and
stage-guided dynamic multi-sensory fusion reweights vision, audio, and
touch according to predicted interaction stages~\cite{Feng2024MSBot}.
These designs show that phase-aware priors help, but they rely on
task-specific assumptions or explicit stage labels and therefore do not
constitute a general, label-free mechanism for contact-rich fusion. SO-TA
instead derives its phase-awareness implicitly, through marginal
constraints on a force--pose-conditioned transport plan.

\paragraph*{Optimal transport in attention}
Entropy-regularized OT~\cite{cuturi2013sinkhorn,peyre2019computational}
admits efficient Sinkhorn solvers and has been used to replace softmax
normalization in attention layers~\cite{Sander2022Sinkformers}, to
sharpen slot-attention assignments~\cite{Zhang2023UnlockingSlotAttention},
and for multi-prompt zero-shot
segmentation~\cite{Kim2024OTSeg}. We bring this OT view of attention to
the visuo-haptic fusion problem and tie its row marginal to a learned
supply distribution that is itself conditioned on force and pose tokens,
yielding a label-free structured prior on the spatial selection step.

\section{Preliminaries}
\label{sec:prelim}
We collect the notation, the diffusion-policy backbone, and the entropic
OT formulation used in the rest of the paper.

\subsection{Tri-modal imitation learning}
\label{subsec:prelim_il}
At each time index $t$ the robot collects synchronized observations
$(\mathbf{o}^{\mathrm{img}}_{t},\mathbf{o}^{\mathrm{force}}_{t},\mathbf{o}^{\mathrm{pose}}_{t})$.
A fixed-length window of length $T_w$ is mapped to a future
pose-action chunk of horizon $T_h$, yielding training tuples
\begin{equation}
\bigl(
\mathbf{o}^{\mathrm{img}}_{t-T_w:t-1},
\mathbf{o}^{\mathrm{force}}_{t-T_w:t-1},
\mathbf{o}^{\mathrm{pose}}_{t-T_w:t-1},
\mathbf{a}^{\mathrm{pose}}_{t:t+T_h-1}
\bigr)\in\mathcal{D}.
\label{eq:windowchunk}
\end{equation}
A behavior-cloning objective
\begin{equation}
\min\; \mathbb{E}_{\mathcal{D}}
\Bigl[\Vert\hat{\mathbf{a}}^{\mathrm{pose}}_{t:t+T_h-1}
-\mathbf{a}^{\mathrm{pose}}_{t:t+T_h-1}\Vert_2^2\Bigr]
\label{eq:bc_obj}
\end{equation}
is a standard baseline but is susceptible to covariate shift; we use it
as a starting point and replace the regressor by a diffusion denoiser.

\subsection{Diffusion-based sequence policy}
\label{subsec:prelim_dp}
Diffusion models \cite{ho2020ddpm,song2021scorebased} learn to invert a
gradual Gaussian noising process
\begin{equation}
\mathbf{y}_{s}=\sqrt{1-\mathcal{G}^{\mathrm{diff}}_{s}}\,\mathbf{y}_{s-1}
+\sqrt{\mathcal{G}^{\mathrm{diff}}_{s}}\,\boldsymbol{\epsilon}_{s},
\;\;\boldsymbol{\epsilon}_{s}\!\sim\!\mathcal{N}(\mathbf{0},\mathbb{I}),
\label{eq:diff_forward}
\end{equation}
$s=1,\ldots,N_{\mathrm{diff}}$, with a fixed schedule
$\mathcal{G}^{\mathrm{diff}}_{s}\!\in\!(0,1)$. A conditional denoiser is
trained with the $\epsilon$-prediction objective
\begin{equation}
\min\; \mathbb{E}_{s,\mathbf{y}_0,\boldsymbol{\epsilon}_s}
\Bigl[\Vert\boldsymbol{\epsilon}_s-\widehat{\boldsymbol{\epsilon}}(\mathbf{y}_s,s,\mathbf{z}^{\mathrm{cond}})\Vert_2^2\Bigr].
\label{eq:diff_eps}
\end{equation}
Following Diffusion Policy~\cite{Chi-RSS-23} we set
\begin{equation}
\mathbf{y}_0\triangleq\mathbf{a}^{\mathrm{pose}}_{t:t+T_h-1},\qquad
\mathbf{z}^{\mathrm{cond}}\triangleq\mathbf{Z}^{\mathrm{fused}}_{t-T_w:t-1}.
\end{equation}
Online, an $N^{\mathrm{infer}}_{\mathrm{diff}}\!\ll\!N_{\mathrm{diff}}$-step
reverse process yields a low-latency receding-horizon controller.

\subsection{Entropy-regularized Optimal Transport}
\label{subsec:prelim_ot}
Given a cost matrix
$\mathcal{C}\!\in\!\Real^{N_{\mathrm{sub}}\times N_{\mathrm{patch}}}$,
a supply $\boldsymbol{\gamma}\!\in\!\Real^{N_{\mathrm{sub}}}$, and a
capacity $\boldsymbol{\beta}\!\in\!\Real^{N_{\mathrm{patch}}}$, the
entropy-regularized OT problem solves
\begin{align}
\min_{\mathbf{\Pi}\ge0}\;
&\langle\mathbf{\Pi},\mathcal{C}\rangle
-\varepsilon_{\mathrm{ot}}H(\mathbf{\Pi})\label{eq:ot_obj}\\
\text{s.t.}\;&
\mathbf{\Pi}\mathbf{1}=\boldsymbol{\gamma},\;
\mathbf{\Pi}^{\mathsf T}\mathbf{1}=\boldsymbol{\beta},
\label{eq:ot_marginals}
\end{align}
with entropy $H(\mathbf{\Pi})=-\sum_{\ell,p}\Pi_{\ell,p}\log\Pi_{\ell,p}$
and temperature $\varepsilon_{\mathrm{ot}}\!>\!0$. The Sinkhorn
algorithm~\cite{cuturi2013sinkhorn} solves
\eqref{eq:ot_obj}--\eqref{eq:ot_marginals} via alternating log-domain
updates; we use this formulation throughout
Sec.~\ref{sec:sota}.

\section{Problem Formulation}
\label{sec:problem}
\begin{figure}[!t]
\centering
\includegraphics[width=0.95\columnwidth]{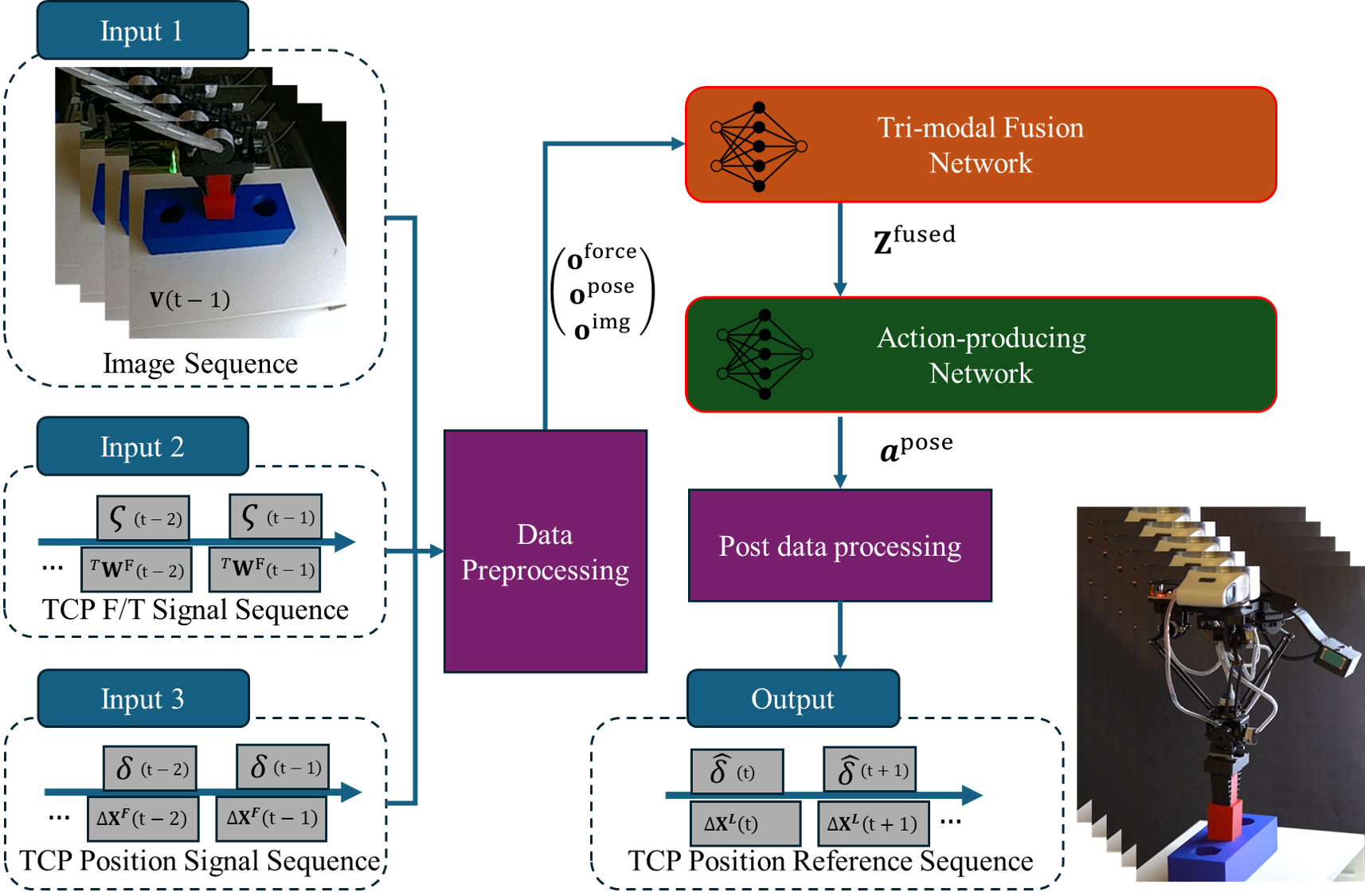}
\caption{Overall forward pass of the tri-modal imitation-learning
pipeline: an observation window is preprocessed, fused into a single
encoding, and consumed by a diffusion policy that emits a pose-action
chunk; the chunk is integrated and clamped before being sent to an
impedance loop.}
\label{fig:pipeline}
\end{figure}

\paragraph*{Pipeline}
Figure~\ref{fig:pipeline} summarizes the data flow. At inference cadence
$f_{\mathrm{infer}}$, the follower-side sensors expose an RGB
sequence~$\mathbf{V}$, the TCP wrench~${}^{T}\mathbf{W}^{F}$ together
with a gripper force scalar~$\varsigma$, and the TCP incremental
motion~$\Delta\mathbf{X}^{F}$ together with the gripper
position~$\delta$. Preprocessing time-aligns the streams, segments them
into windows of length~$T_w$, converts the pose channel into
frame-to-frame increments, and normalizes low-dimensional channels with
dataset statistics. Each RGB frame is center-cropped, resized, and
normalized in pixel space.

\paragraph*{Why a structured fusion module matters}
Contact-rich tasks display a phase-dependent dominance of
modalities: vision concentrates relevant cues in a small spatial region
before contact, while force/torque becomes the primary signal during
contact regulation. A fusion module that reflects this regularity acts
as an inductive bias and improves data efficiency. SO-TA encodes this
prior by replacing softmax with an OT alignment whose row marginal is
conditioned on the force--pose token, biasing spatial selection toward
patches that are jointly consistent with the current haptic state.

\section{Spacetime Optimal-Transport Attention}
\label{sec:sota}

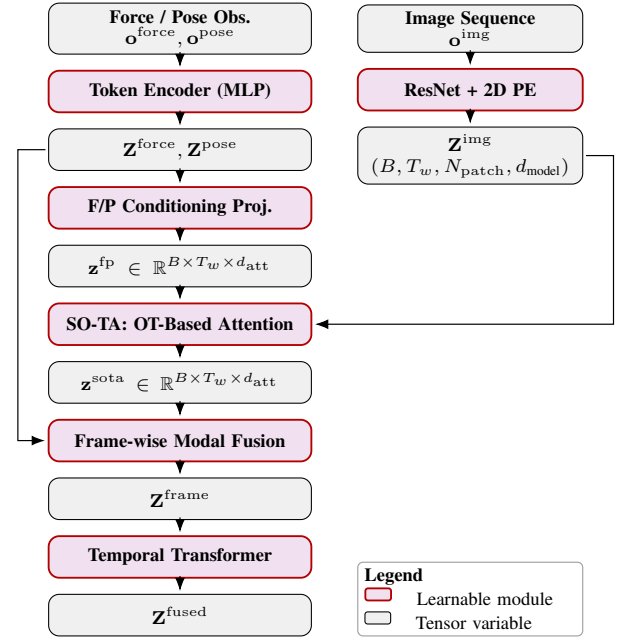
\begin{figure}[!t]
\centering
\begin{tikzpicture}[
  font=\scriptsize,
  >=Latex,
  node distance=2.0mm and 4.0mm,
  procfill/.style={fill=violet!12},
  varfill/.style={fill=gray!12},
  varbox/.style={draw, rounded corners, varfill,
                 align=center, inner sep=2.5pt,
                 text width=33mm, minimum height=5.5mm},
  procbox/.style={draw, rounded corners, procfill,
                  align=center, inner sep=2.5pt,
                  text width=33mm, minimum height=5.5mm},
  imgbox/.style={draw, rounded corners, varfill,
                 align=center, inner sep=2.5pt,
                 text width=28mm, minimum height=5.5mm},
  imgproc/.style={draw, rounded corners, procfill,
                  align=center, inner sep=2.5pt,
                  text width=28mm, minimum height=5.5mm},
  learn/.style={draw=red!70!black, line width=0.7pt},
  flow/.style={->, line width=0.45pt, shorten <=1pt, shorten >=1pt},
  legendbox/.style={draw=black!40, rounded corners=2pt, fill=white,
                    align=left, inner sep=2.5pt, text width=28mm},
]
\node[varbox] (in)
   {\textbf{Force / Pose Obs.}\\
   $\mathbf{o}^{\mathrm{force}},\mathbf{o}^{\mathrm{pose}}$};
\node[procbox, learn, below=of in] (tok)
   {\textbf{Token Encoder (MLP)}};
\node[varbox, below=of tok] (tokv)
   {$\mathbf{Z}^{\mathrm{force}},\mathbf{Z}^{\mathrm{pose}}$};
\node[procbox, learn, below=of tokv] (fp)
   {\textbf{F/P Conditioning Proj.}};
\node[varbox, below=of fp] (fpv)
   {$\mathbf{z}^{\mathrm{fp}}\in\Real^{B\times T_w\times d_{\mathrm{att}}}$};
\node[procbox, learn, below=of fpv] (sota)
   {\textbf{SO-TA: OT-Based Attention}};
\node[varbox, below=of sota] (zsota)
   {$\mathbf{z}^{\mathrm{sota}}\in\Real^{B\times T_w\times d_{\mathrm{att}}}$};
\node[procbox, learn, below=of zsota] (frame)
   {\textbf{Frame-wise Modal Fusion}};
\node[varbox, below=of frame] (zframe)
   {$\mathbf{Z}^{\mathrm{frame}}$};
\node[procbox, learn, below=of zframe] (temp)
   {\textbf{Temporal Transformer}};
\node[varbox, below=of temp] (zout)
   {$\mathbf{Z}^{\mathrm{fused}}$};

\node[imgbox, right=6mm of in] (img)
   {\textbf{Image Sequence}\\
   $\mathbf{o}^{\mathrm{img}}$};
\node[imgproc, learn, below=of img] (enc)
   {\textbf{ResNet + 2D PE}};
\node[imgbox, below=of enc] (zimg)
   {$\mathbf{Z}^{\mathrm{img}}$\\
   $(B,T_w,N_{\mathrm{patch}},d_{\text{model}})$};

\draw[flow] (in) -- (tok);
\draw[flow] (tok) -- (tokv);
\draw[flow] (tokv) -- (fp);
\draw[flow] (fp) -- (fpv);
\draw[flow] (fpv) -- (sota);
\draw[flow] (sota) -- (zsota);
\draw[flow] (zsota) -- (frame);
\draw[flow] (frame) -- (zframe);
\draw[flow] (zframe) -- (temp);
\draw[flow] (temp) -- (zout);
\draw[flow] (img) -- (enc);
\draw[flow] (enc) -- (zimg);
\path (img.east) ++(4mm,0) coordinate (RB);
\draw[flow] (zimg.east) -- (RB |- zimg.east) -- (RB |- sota.east) -- (sota.east);
\path (in.west) ++(-4mm,0) coordinate (LB);
\draw[flow] (tokv.west) -- (LB |- tokv.west) -- (LB |- frame.west) -- (frame.west);

\path (zimg.east |- zout.south) coordinate (LegendAnchor);
\node[legendbox, anchor=south east] at (LegendAnchor) {%
\textbf{Legend}\\[0.3mm]
\tikz[baseline=-0.6ex]\draw[draw=red!70!black, line width=0.7pt, fill=violet!12, rounded corners=1pt] (0,0) rectangle (0.34,0.16);
\quad Learnable module\\[0.3mm]
\tikz[baseline=-0.6ex]\draw[draw=black, fill=gray!12, rounded corners=1pt] (0,0) rectangle (0.34,0.16);
\quad Tensor variable
};
\end{tikzpicture}
\caption{End-to-end SO-TA fusion pipeline. Red-bordered blocks are
learnable, gray blocks are tensor variables. The OT-based attention is
detailed in Fig.~\ref{fig:sota_compact}.}
\label{fig:sota_pipeline}
\end{figure}

Figure~\ref{fig:sota_pipeline} summarizes the SO-TA backbone over a window
of length $T_w$. The goal is a time-aligned fused encoding
$\mathbf{Z}^{\mathrm{fused}}\!\in\!\Real^{B\times T_w\times d_{\text{model}}}$.

\subsection{Token encoding and force--pose conditioning}
Force and pose observations are mapped by learnable MLP encoders into
token sequences
$\mathbf{Z}^{\mathrm{force}},\mathbf{Z}^{\mathrm{pose}}
\!\in\!\Real^{B\times T_w\times d_{\text{model}}}$ and concatenated and
projected to form the conditioning embedding
$\mathbf{z}^{\mathrm{fp}}\!\in\!\Real^{B\times T_w\times d_{\mathrm{att}}}$.
Each RGB frame is encoded by a ResNet-18~\cite{he2016deep} with
GroupNorm into patch tokens
$\mathbf{Z}^{\mathrm{img}}\!\in\!\Real^{B\times T_w\times N_{\mathrm{patch}}\times d_{\text{model}}}$
augmented with a deterministic 2D positional encoding.

\subsection{OT-based attention}
\label{subsec:sota_ot}
\begin{figure}[!t]
\centering
\includegraphics[width=0.98\columnwidth]{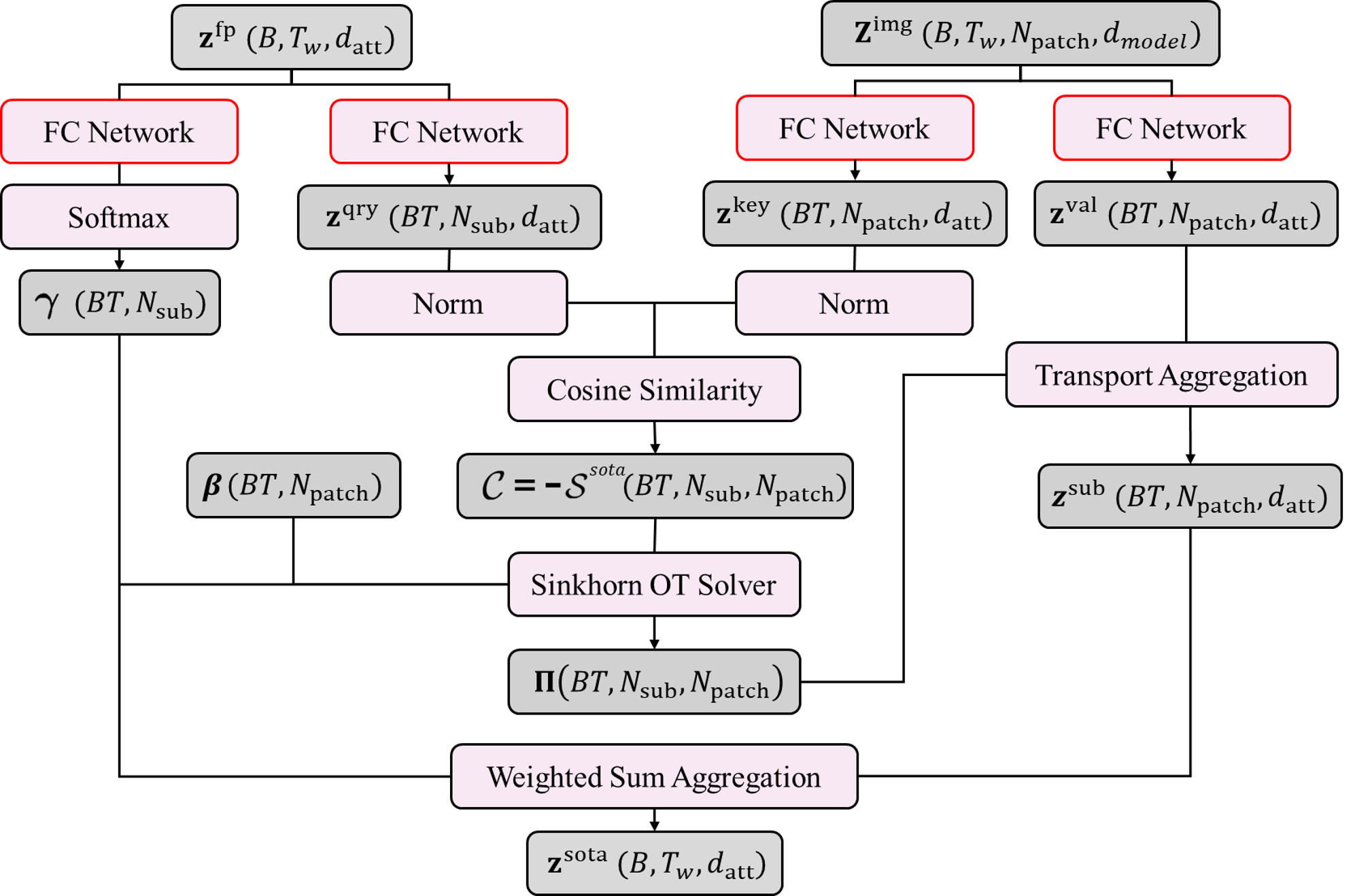}
\caption{Compact view of the OT-based attention node inside SO-TA. The
sub-queries and the row-marginal supply~$\boldsymbol{\gamma}$ are both
generated from the force--pose conditioning vector; the column marginal
is uniform; transport is solved by log-domain Sinkhorn.}
\label{fig:sota_compact}
\end{figure}

For notational convenience we merge $(B,T_w)$ into $BT$ and index a
frame sample by $n$, sub-queries by $\ell$, and patches by $p$.

\textbf{1) Sub-query and supply.}
A FC head produces $N_{\mathrm{sub}}$ sub-queries
\begin{equation}
\mathbf{z}^{\mathrm{qry}}_{n}=\operatorname{proj}_{\mathrm{qry}}(\mathbf{z}^{\mathrm{fp}}_n)
\in\Real^{N_{\mathrm{sub}}\times d_{\mathrm{att}}},
\end{equation}
and a second head predicts a row-marginal supply
\begin{equation}
\boldsymbol{\gamma}_n=\operatorname{softmax}\bigl(\operatorname{proj}_{\gamma}(\mathbf{z}^{\mathrm{fp}}_n)\bigr)
\in\Real^{N_{\mathrm{sub}}},\;\sum_{\ell}\gamma_{n,\ell}=1.
\label{eq:gamma}
\end{equation}
Intuitively, $\gamma_{n,\ell}$ is the OT mass carried by sub-query~$\ell$
under the current haptic state.

\textbf{2) Key/value projection.} Each patch is projected to the OT space
\begin{equation}
\mathbf{z}^{\mathrm{key}}_{n,p}=\operatorname{proj}_{\mathrm{key}}(\mathbf{Z}^{\mathrm{img}}_{n,p}),\;
\mathbf{z}^{\mathrm{val}}_{n,p}=\operatorname{proj}_{\mathrm{val}}(\mathbf{Z}^{\mathrm{img}}_{n,p}).
\end{equation}

\textbf{3) Cost.} Sub-queries and Keys are $\ell_2$-normalized; the cost
is the negative cosine similarity
\begin{equation}
\mathcal{C}_{n,\ell,p}=-\bigl\langle\operatorname{norm}(\mathbf{z}^{\mathrm{qry}}_{n,\ell}),
\operatorname{norm}(\mathbf{z}^{\mathrm{key}}_{n,p})\bigr\rangle.
\label{eq:cost}
\end{equation}

\textbf{4) Entropic OT.} With uniform patch capacity
$\boldsymbol{\beta}\!=\!\tfrac{1}{N_{\mathrm{patch}}}\mathbf{1}$, the
transport plan is the solution of
\begin{align}
\min_{\mathbf{\Pi}_n\ge0}\;
&\langle\mathbf{\Pi}_n,\mathcal{C}_n\rangle
-\varepsilon_{\mathrm{ot}}H(\mathbf{\Pi}_n)\nonumber\\
\text{s.t.}\;&
\mathbf{\Pi}_n\mathbf{1}=\boldsymbol{\gamma}_n,\;
\mathbf{\Pi}_n^{\mathsf T}\mathbf{1}=\boldsymbol{\beta}.
\label{eq:sota_ot}
\end{align}
The Sinkhorn solver uses log-domain dual variables
$\boldsymbol{\kappa}_n,\boldsymbol{\nu}_n$ updated by
\begin{align}
\kappa_{n,\ell}&\leftarrow
\log\gamma_{n,\ell}-\log\!\sum_{p}\!\exp\!\Bigl(-\tfrac{\mathcal{C}_{n,\ell,p}}{\varepsilon_{\mathrm{ot}}}+\nu_{n,p}\Bigr),\\
\nu_{n,p}&\leftarrow
\log\beta_{p}-\log\!\sum_{\ell}\!\exp\!\Bigl(-\tfrac{\mathcal{C}_{n,\ell,p}}{\varepsilon_{\mathrm{ot}}}+\kappa_{n,\ell}\Bigr),
\end{align}
and the plan is recovered as
$\Pi_{n,\ell,p}\!=\!\exp(\kappa_{n,\ell}-\mathcal{C}_{n,\ell,p}/\varepsilon_{\mathrm{ot}}+\nu_{n,p})$.

\textbf{5) Aggregation and tied merge.}
Each sub-query collects a message
$\mathbf{z}^{\mathrm{sub}}_{n,\ell}=\sum_p\Pi_{n,\ell,p}\mathbf{z}^{\mathrm{val}}_{n,p}$.
To avoid a degenerate uniform fallback under uniform~$\boldsymbol{\beta}$,
we tie the merge weights to the OT row marginals,
\begin{equation}
\mathbf{z}^{\mathrm{sota}}_{n}=\sum_{\ell}\gamma_{n,\ell}\mathbf{z}^{\mathrm{sub}}_{n,\ell}
=\sum_{p}\zeta^{\mathrm{sota}}_{n,p}\mathbf{z}^{\mathrm{val}}_{n,p},
\label{eq:sota_merge}
\end{equation}
with $\zeta^{\mathrm{sota}}_{n,p}=\sum_{\ell}\gamma_{n,\ell}\Pi_{n,\ell,p}$.

\paragraph*{Remark (why tie $\boldsymbol{\gamma}$).}
Because $\boldsymbol{\beta}$ is uniform, any conditioning-independent
merge over sub-queries collapses to scaled mean pooling over patches and
loses the dependence on $\mathbf{z}^{\mathrm{fp}}_n$. Tying the merge
weights to $\boldsymbol{\gamma}_n$ preserves conditioning while
introducing no additional parameters.

\subsection{Frame-wise and temporal fusion}
At each frame $t$ a length-$3$ token sequence
$[\mathbf{Z}^{\mathrm{force}},\mathbf{Z}^{\mathrm{pose}},
\mathbf{Z}^{\mathrm{sota}}]$ is processed by a lightweight Transformer
shared across frames. A pooled head produces a softmax gate
$\boldsymbol{\mathfrak{w}}_{\mathrm{gate}}\!\in\!\Real^3$ used to combine
the contextualized tokens into a frame-wise fused feature
$\mathbf{Z}^{\mathrm{frame}}$. Sinusoidal time encoding is added and a
temporal Transformer produces the final $\mathbf{Z}^{\mathrm{fused}}$.

\subsection{Interpretability signals}
\paragraph*{Modal influence ratio}
The gate $\boldsymbol{\mathfrak{w}}_{\mathrm{gate}}$ reflects gating over
contextualized tokens and therefore does not isolate raw-modality
contributions. We instead define a leave-one-out ratio: for each
modality $m\!\in\!\{\mathrm{force},\mathrm{pose},\mathrm{sota}\}$ we zero
the corresponding token before the frame-wise Transformer and rerun
fusion to obtain $\mathbf{Z}^{\mathrm{frame},m}_{t}$. The ratio
\begin{equation}
\mathfrak{w}^{m}_{t}=
\frac{\Vert\mathbf{Z}^{\mathrm{frame}}_{t}-\mathbf{Z}^{\mathrm{frame},m}_{t}\Vert_2}
{\sum_{m'}\Vert\mathbf{Z}^{\mathrm{frame}}_{t}-\mathbf{Z}^{\mathrm{frame},m'}_{t}\Vert_2}
\label{eq:modal_ratio}
\end{equation}
is non-negative and sums to one across modalities.

\paragraph*{OT-derived heatmap}
We scale the normalized effective patch weights by the visual influence
ratio,
\begin{equation}
\zeta^{\mathrm{heatmap}}_{n,p}
=\mathfrak{w}^{\mathrm{sota}}_{n}\cdot
\frac{\zeta^{\mathrm{sota}}_{n,p}}{\sum_{p'}\zeta^{\mathrm{sota}}_{n,p'}},
\label{eq:heatmap}
\end{equation}
and reshape into the patch grid for overlay.

\section{Baselines}
\label{sec:baselines}
\subsection{Concatenation}
A learnable MLP encodes the force observation into
$\mathbf{z}^{\mathrm{force}}\!\in\!\Real^{B\times T_w\times d_{\mathrm{forcefeat}}}$
and a ResNet encodes each RGB frame into
$\mathbf{z}^{\mathrm{vis}}\!\in\!\Real^{B\times T_w\times d_{\mathrm{vis}}}$.
The pose observation is appended directly. The fused tensor is
\begin{equation}
\mathbf{Z}^{\mathrm{fused}}=
[\mathbf{z}^{\mathrm{vis}};\,\mathbf{z}^{\mathrm{force}};\,
\mathbf{o}^{\mathrm{pose}}]
\in\Real^{B\times T_w\times d_{\mathrm{cat}}},
\end{equation}
with $d_{\mathrm{cat}}=d_{\mathrm{vis}}+d_{\mathrm{forcefeat}}+d_{\mathrm{pose}}$.
Cross-modal interaction must be learned downstream.

\subsection{Force--pose-conditioned cross-attention}
The module shares the interfaces of Fig.~\ref{fig:sota_pipeline} but
replaces the OT step with standard scaled dot-product attention
\cite{Vaswani2017}. A single query
$\mathbf{z}^{\mathrm{qry}}_{n}=\operatorname{proj}_{\mathrm{qry}}(\mathbf{z}^{\mathrm{fp}}_n)$
is matched against patch Keys; weights
$\zeta^{ca}_{n,p}\!=\!\operatorname{softmax}_p
(\langle\operatorname{norm}(\mathbf{z}^{\mathrm{qry}}_{n}),\operatorname{norm}(\mathbf{z}^{\mathrm{key}}_{n,p})\rangle/\sqrt{d_{\mathrm{att}}})$
aggregate Values into
$\mathbf{z}^{\mathrm{ca}}_{n}=\sum_{p}\zeta^{ca}_{n,p}\mathbf{z}^{\mathrm{val}}_{n,p}$.
Mixed-precision-stable normalization follows
\cite{micikevicius2018mixedprecision}.

\section{Experiments}
\label{sec:experiments}

\subsection{Implementation and protocol}
\label{subsec:impl}

\begin{table}[!t]
\caption{Key Hyperparameters Used in All Experiments\label{tab:params}}
\centering
\renewcommand{\arraystretch}{1.05}
\setlength{\tabcolsep}{4pt}
\small
\begin{tabular}{@{}lll@{}}
\toprule
\textbf{Symbol} & \textbf{Value} & \textbf{Meaning} \\
\midrule
$T_w$ & 8 & Observation window length \\
$T_h$ & 16 & Action horizon (chunk length) \\
$f_{\mathrm{infer}}$ & 10\,Hz & Online inference frequency \\
$B$ & 256 & Training batch size \\
$H_{\mathrm{img}}\!\times\!W_{\mathrm{img}}$ & $224\!\times\!224$ & Image resolution \\
$d_{\mathrm{force}},d_{\mathrm{pose}}$ & 9,\,10 & Force / pose obs.\ dim. \\
$N_{\mathrm{diff}}$ & 100 & Training diffusion steps \\
$N^{\mathrm{infer}}_{\mathrm{diff}}$ & 10 & Inference denoising steps \\
$d_{\text{model}}$ & 512 & Backbone feature dim. \\
$d_{\mathrm{att}}$ & 256 & OT / attention dim. \\
$h\!\times\!w\!=\!N_{\mathrm{patch}}$ & $7\!\times\!7\!=\!49$ & Patch grid \\
$N_{\mathrm{sub}}$ & 2 & OT sub-queries \\
$\varepsilon_{\mathrm{ot}}$ & 0.20 & Entropic OT temperature \\
$N_{\mathrm{ot}}$ & 5 & Sinkhorn iterations \\
$\boldsymbol{\beta}$ & $\tfrac{1}{49}\mathbf{1}$ & Patch capacity \\
$n_h$ & 4 & Attention heads (CA) \\
\bottomrule
\end{tabular}
\end{table}

Demonstrations are collected with a leader--follower bilateral
teleoperation setup~\cite{feng2026delta6,feng2026one,feng2024wos}: the
teleoperator perceives the scene only through the wrist-mounted RGB
stream and the leader-side haptic feedback, matching the sensing
available to the policy at deployment. Only follower-side signals are
logged. Force and pose channels are scale-stabilized
(magnitude/direction split for F/T; 6D rotation representation
\cite{zhou2019continuity}) and normalized with dataset min--max
statistics. Each RGB frame is center-cropped, resized to $224\times224$
and normalized. The diffusion denoiser is a 1D conditional UNet over the
temporal axis of the action chunk; we use a squared-cosine schedule,
AdamW with cosine warm-up, mixed-precision training, and EMA weights for
evaluation. Hyperparameters are summarized in
Table~\ref{tab:params}. Training runs on a single NVIDIA RTX~6000~Pro
for $4{,}000$ epochs with $8{,}192$ samples per epoch ($\sim$22~h).
Inference is benchmarked on an RTX~4070~Ti~SUPER.

\begin{table}[!t]
\caption{Parameter Counts of the Three Fusion Backbones\label{tab:params_count}}
\centering
\renewcommand{\arraystretch}{1.05}
\small
\begin{tabular}{lrrr}
\toprule
Fusion type & DP params & Fusion params & Total \\
\midrule
SO-TA            & 40\,M & 21\,M & 61\,M \\
Cross-attention & 40\,M & 21\,M & 61\,M \\
Concatenation   & 69\,M & 0\,M  & 81\,M \\
\bottomrule
\end{tabular}
\end{table}

\subsection{Tight peg-in-hole assembly}
\label{subsec:peg}

\begin{figure}[!t]
\centering
\includegraphics[width=0.95\columnwidth]{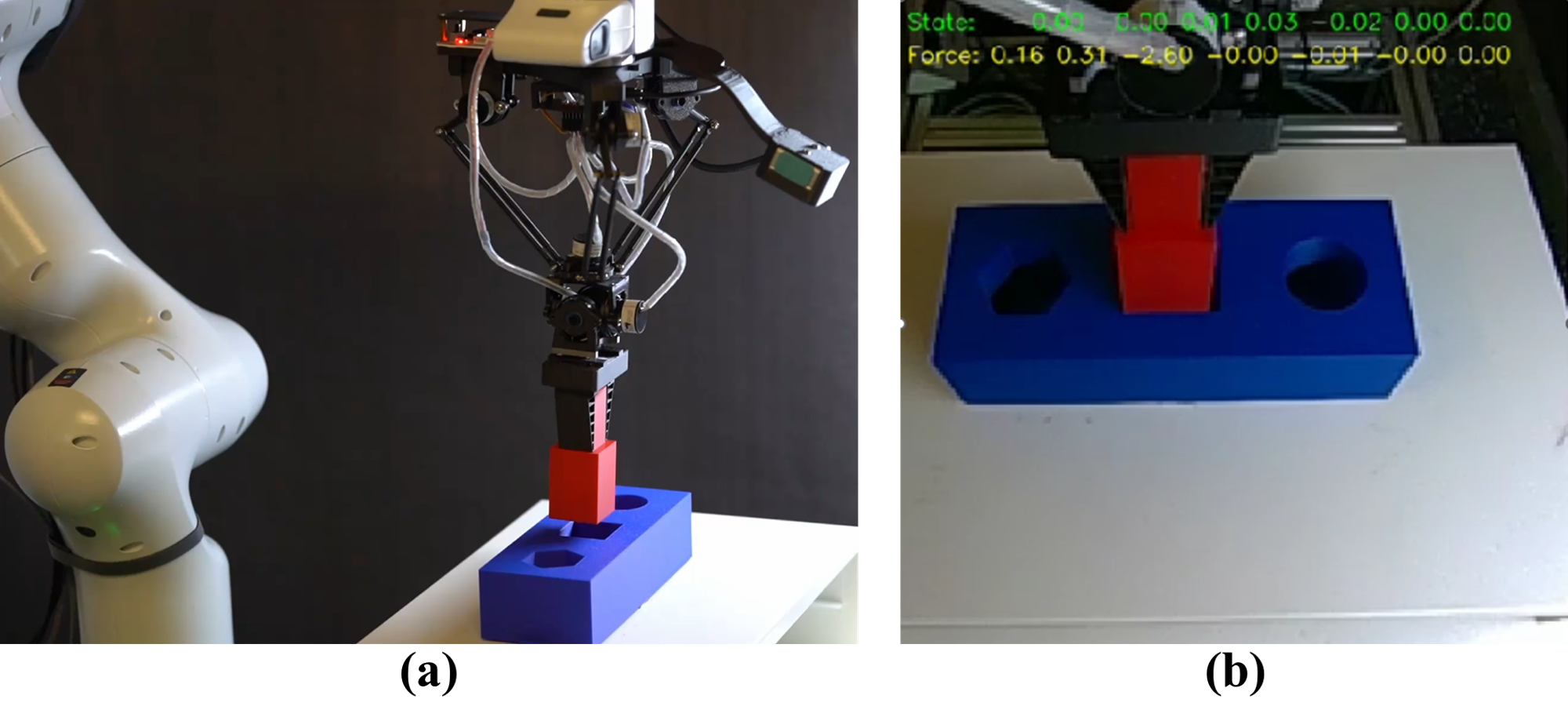}
\caption{Tight peg-in-hole task. (a) Third-person view of the setup.
(b) Robot-mounted camera view, i.e.\ the visual observation seen by the
policy.}
\label{fig:peg_setup}
\end{figure}

The peg-in-hole testbed (Fig.~\ref{fig:peg_setup}) has clearance below
$0.2~\mathrm{mm}$, which makes contact dynamics critical. We collect
$198$ teleoperated episodes ($32{,}974$ synchronized tri-modal steps;
$3{,}297.4~\mathrm{s}$). With $T_w\!=\!8$, $T_h\!=\!16$ and episode-boundary
padding, this yields $29{,}608$ window--chunk pairs. The initial pose is
randomized within bounded translation/rotation perturbations.

\subsubsection{Inference latency vs.\ denoising iterations}
\begin{table}[!t]
\caption{Inference Latency and First-Step $z$ MAE vs.\ $N^{\mathrm{infer}}_{\mathrm{diff}}$ for the SO-TA Policy\label{tab:peg_timing}}
\centering
\renewcommand{\arraystretch}{1.05}
\setlength{\tabcolsep}{4pt}
\resizebox{\columnwidth}{!}{%
\begin{tabular}{@{}lrrrrrrrrrr@{}}
\toprule
$N^{\mathrm{infer}}_{\mathrm{diff}}$ & 1 & 2 & 3 & 4 & 5 & 6 & 7 & 8 & 9 & 10\\
\midrule
Avg inference time (ms) & 9.6 & 8.5 & 9.5 & 10.5 & 11.3 & 12.3 & 13.3 & 14.2 & 15.2 & 16.2\\
Equiv.\ max $f_{\mathrm{infer}}$ (Hz) & 105 & 118 & 105 & 96 & 89 & 82 & 75 & 70 & 66 & 62\\
$z$ MAE $\times10^{-3}$ (m) & 5.37 & 0.14 & 0.09 & 0.09 & 0.08 & 0.08 & 0.08 & 0.09 & 0.08 & 0.08\\
\bottomrule
\end{tabular}%
}
\end{table}

\begin{figure}[!t]
\centering
\includegraphics[width=0.98\columnwidth]{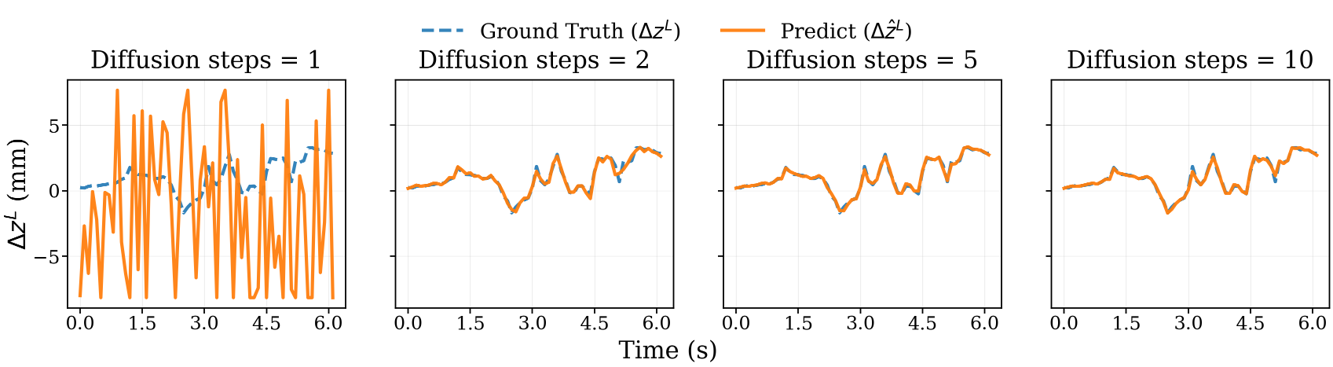}
\caption{Predicted first-step $\Delta z^{L}$ vs.\ ground truth for
$N^{\mathrm{infer}}_{\mathrm{diff}}\!\in\!\{1,2,5,10\}$ (SO-TA).}
\label{fig:peg_diff}
\end{figure}

A control deadline of $1/f_{\mathrm{infer}}\!=\!100~\mathrm{ms}$ at
$10~\mathrm{Hz}$ is the binding constraint. With
$N^{\mathrm{infer}}_{\mathrm{diff}}\!=\!1$ the reverse process starts from
high noise and the prediction is dominated by noise; after the second
iteration the MAE drops sharply and the predicted trajectory closely
tracks ground truth (Fig.~\ref{fig:peg_diff} and
Table~\ref{tab:peg_timing}). Beyond $N^{\mathrm{infer}}_{\mathrm{diff}}\!=\!3$
gains are marginal, and even $N^{\mathrm{infer}}_{\mathrm{diff}}\!=\!10$
fits comfortably inside the deadline.

\subsubsection{Modality ablation}
\begin{figure}[!t]
\centering
\includegraphics[width=0.98\columnwidth]{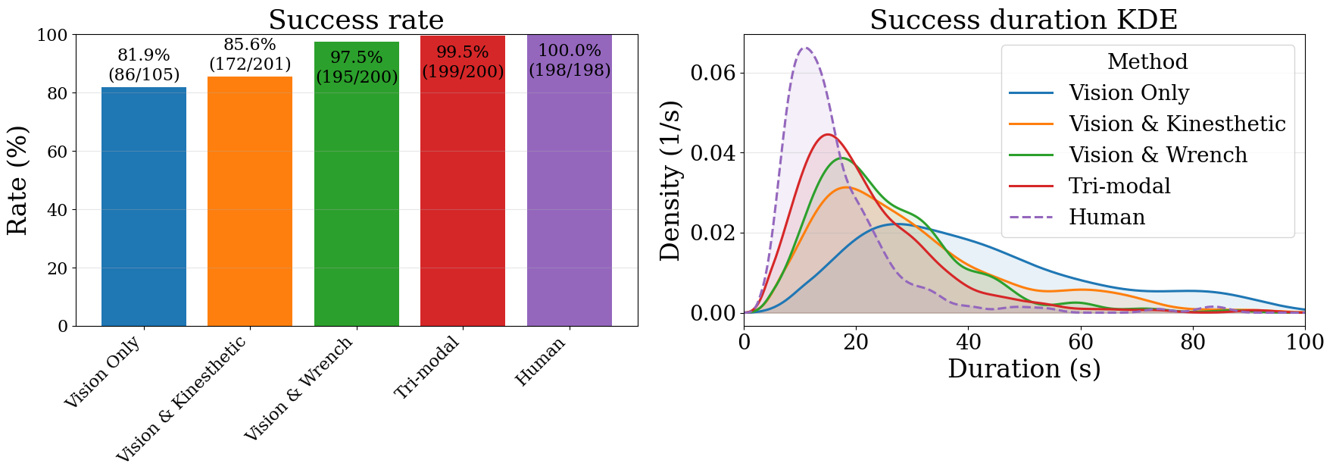}
\caption{Peg-in-hole: success rate (left) and success-only completion
time density (KDE, right) under modality zero-masking with the
concatenation backbone.}
\label{fig:peg_ablation}
\end{figure}

Holding the concatenation backbone fixed, we ablate F/T and pose by
zero-masking the corresponding input tensors during both training and
rollouts. Each variant runs $\sim\!200$ rollouts under a $5~\mathrm{N}$
safety threshold and a $100~\mathrm{s}$ timeout. The full tri-modal input
achieves the highest success ($\sim\!99.5\%$) and the most concentrated
duration density (Fig.~\ref{fig:peg_ablation}). Removing pose has a mild
effect ($97.5\%$ success); removing F/T drops success to $85.6\%$ and
spreads completion time. Vision-only attains the lowest success
($81.9\%$). Failures with zero-masked F/T are dominated by exceeding the
safety threshold; with F/T available, failures more often follow a
covariate-shift drift pattern. All three modalities contribute; F/T has
the strongest marginal influence after vision.

\subsubsection{Fusion-method comparison}
\begin{figure}[!t]
\centering
\includegraphics[width=0.98\columnwidth]{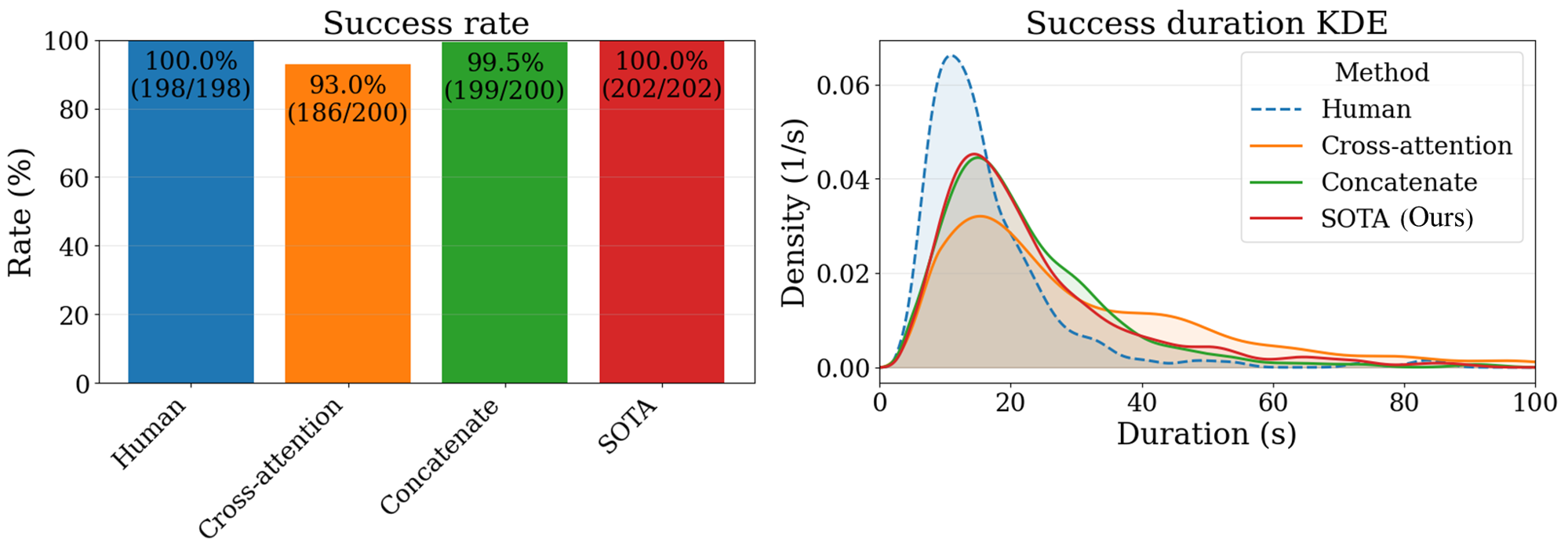}
\caption{Peg-in-hole: success rate (left) and success-only completion
time density (right) across SO-TA, cross-attention, and concatenation.}
\label{fig:peg_compare}
\end{figure}

Holding the input modalities fixed to the full tri-modal set, only the
fusion module varies. SO-TA reaches $100\%$ success across
$\sim\!200$ rollouts; concatenation reaches $99.5\%$ with one failure;
cross-attention drops to $93\%$ (Fig.~\ref{fig:peg_compare}).
Cross-attention also exhibits the heaviest completion-time tail. At
matched parameter budget (Table~\ref{tab:params_count}), SO-TA's explicit
marginal constraints yield more stable spatial selection than
softmax-normalized attention.

\subsubsection{Generalization under visual perturbation}
\begin{figure}[!t]
\centering
\includegraphics[width=0.98\columnwidth]{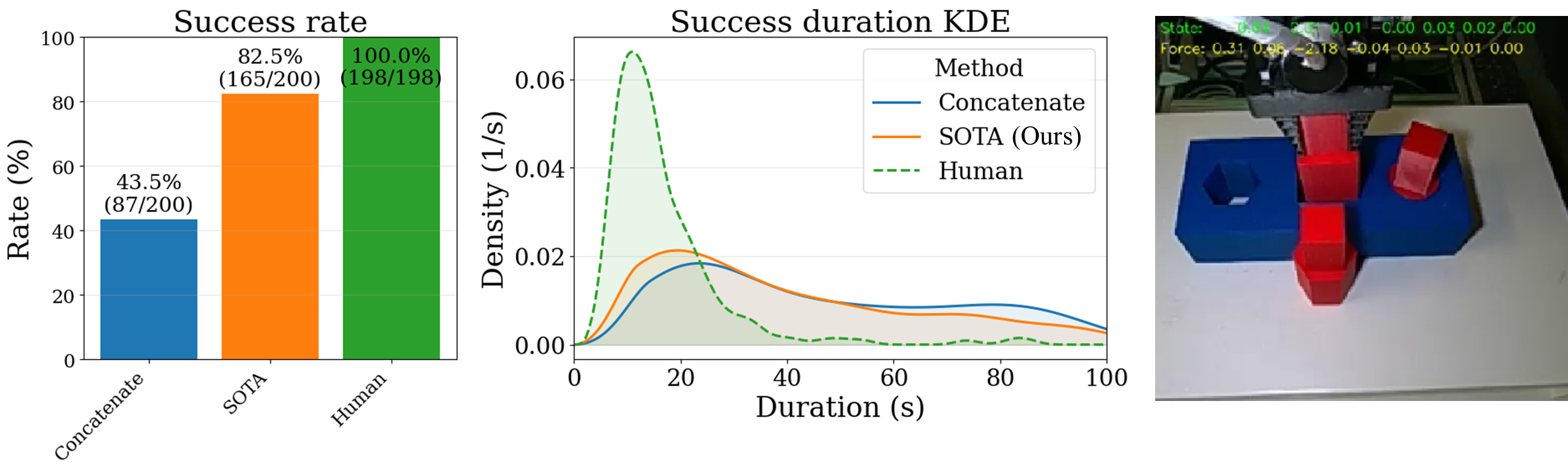}
\caption{Peg-in-hole under illumination changes, distractor pegs, and
partial occlusion. SO-TA retains $82.5\%$ success; concatenation drops to
$43.5\%$.}
\label{fig:peg_gen}
\end{figure}

Without any retraining, we vary the illumination, add distractor pegs,
and partially occlude the socket. Over $200$ rollouts per method, SO-TA
retains $82.5\%$ success ($165/200$); concatenation drops to $43.5\%$
($87/200$) (Fig.~\ref{fig:peg_gen}). Both methods slow down, but SO-TA
retains a clear advantage in success-only completion time. Structured
tri-modal fusion meaningfully buys robustness to visual distribution
shift.

\subsubsection{Interpretability}
\begin{figure}[!t]
\centering
\includegraphics[width=0.98\columnwidth]{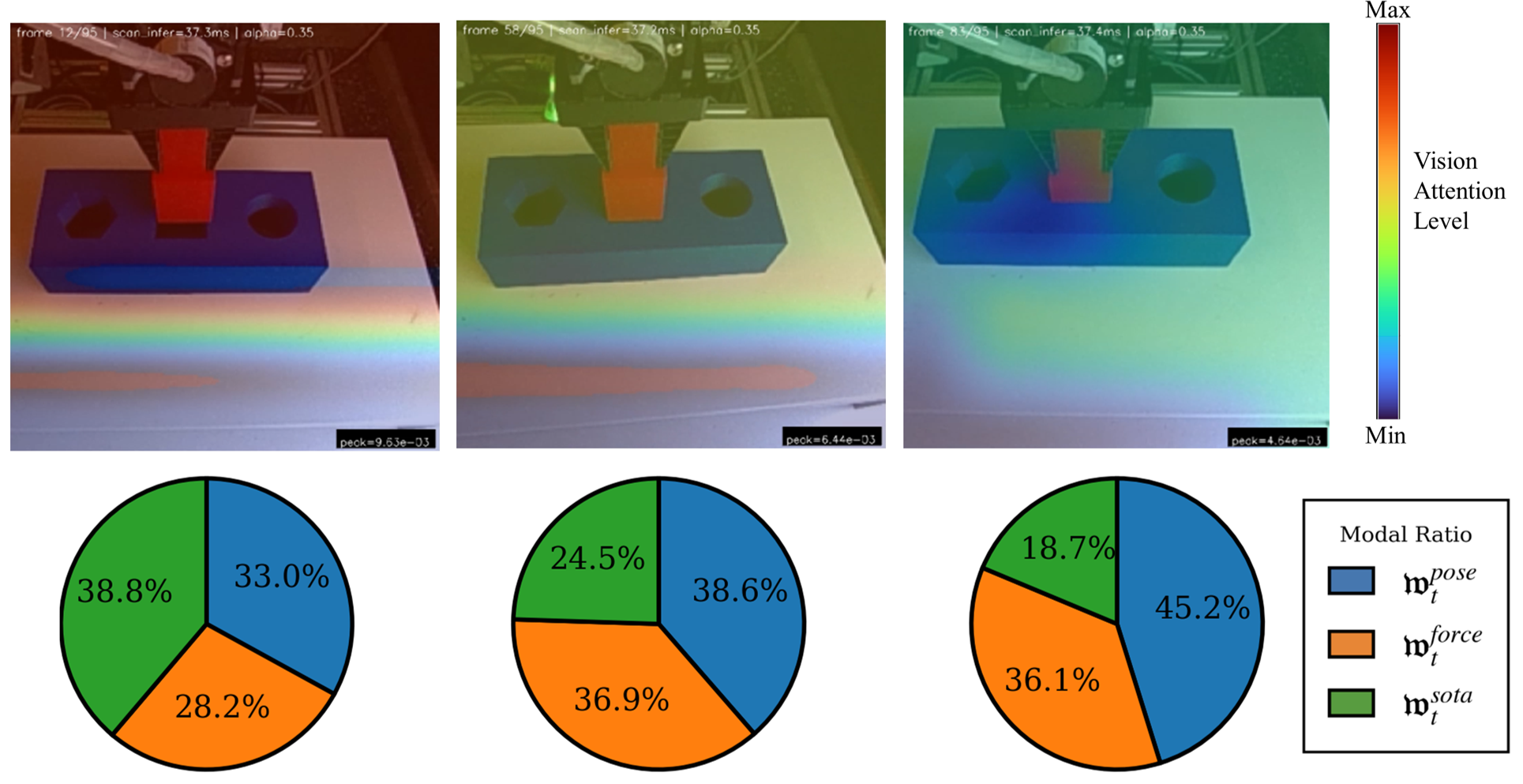}
\caption{Peg-in-hole: SO-TA interpretability across three phases. Top:
OT-based patch heatmap from \eqref{eq:heatmap}. Bottom: modality ratios
\eqref{eq:modal_ratio}. Vision share peaks before contact; F/T and pose
grow during search and insertion.}
\label{fig:peg_explain}
\end{figure}

Replaying the trained SO-TA policy on a demonstration, we report the
RGB observation, the OT heatmap, and the modality ratios at three
representative frames (Fig.~\ref{fig:peg_explain}). During the
pre-contact approach the visual share reaches its maximum across the
trajectory ($38.8\%$), with pose at $33.0\%$ and force at $28.2\%$.
After contact the visual share drops to $24.5\%$ during hole searching
and to $18.7\%$ during insertion, while pose grows to $45.2\%$ and
force settles at $36.1\%$. This matches the expected post-contact
regime, where F/T and pose carry more information than vision for fine
alignment. The heatmaps favor visually distinctive regions; the
diffuse-but-plausible focus is consistent with the absence of an
explicit spatial prior in the OT cost.

\subsection{BCM wiring-connector insertion}
\label{subsec:bcm}

\begin{figure}[!t]
\centering
\includegraphics[width=0.95\columnwidth]{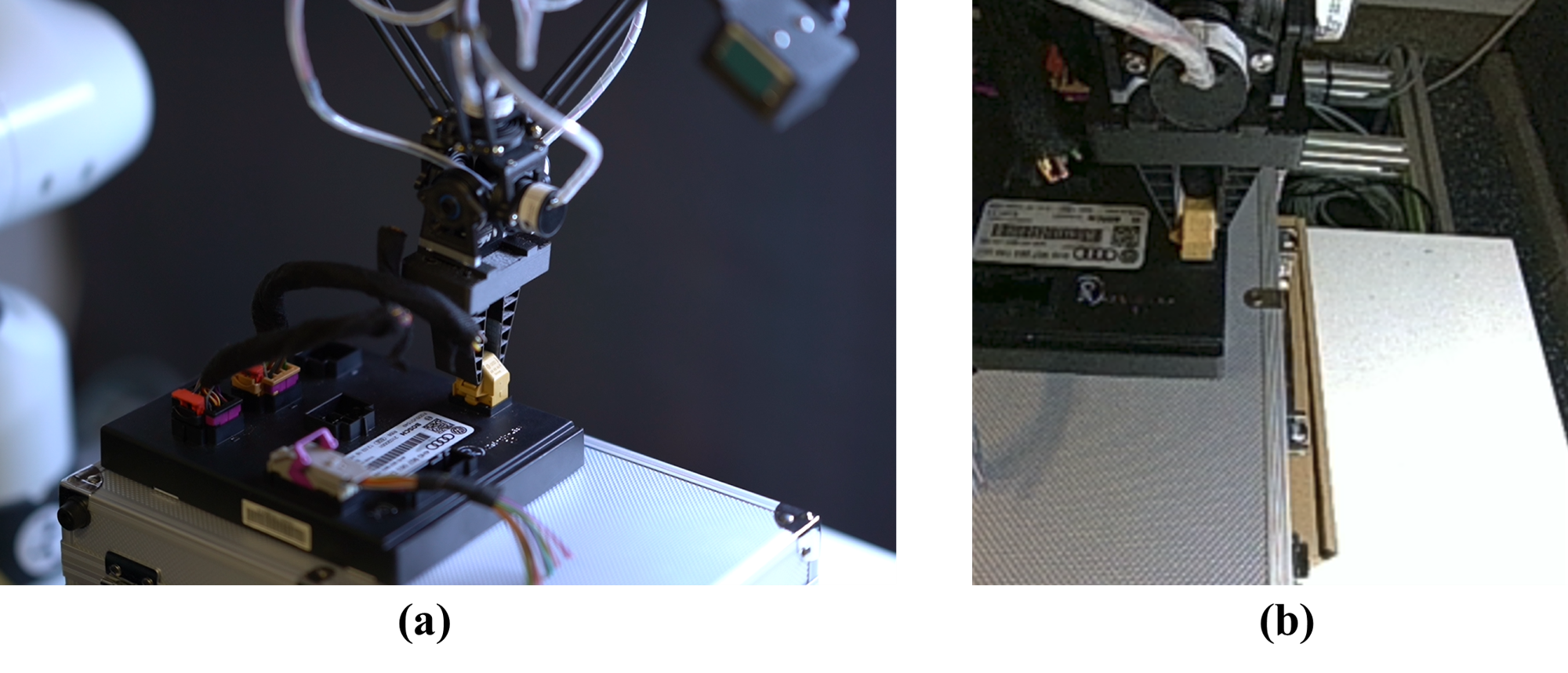}
\caption{BCM wiring-connector insertion. (a) Setup. (b) Robot-camera view.}
\label{fig:bcm_setup}
\end{figure}

The Audi BCM connector task (Fig.~\ref{fig:bcm_setup}) is visually more
challenging than peg-in-hole: the task-relevant region occupies a
smaller fraction of the camera frame, the background is more cluttered,
and color contrast is weaker; the clearance is more forgiving. The
dataset contains $396$ teleoperated demonstrations ($29{,}126$ steps,
$2{,}916.2~\mathrm{s}$), yielding $22{,}430$ training pairs at the same
$T_w,T_h$. Success requires the connector to be fully seated under
$5~\mathrm{N}$ downward force.

\begin{figure}[!t]
\centering
\includegraphics[width=0.73\columnwidth]{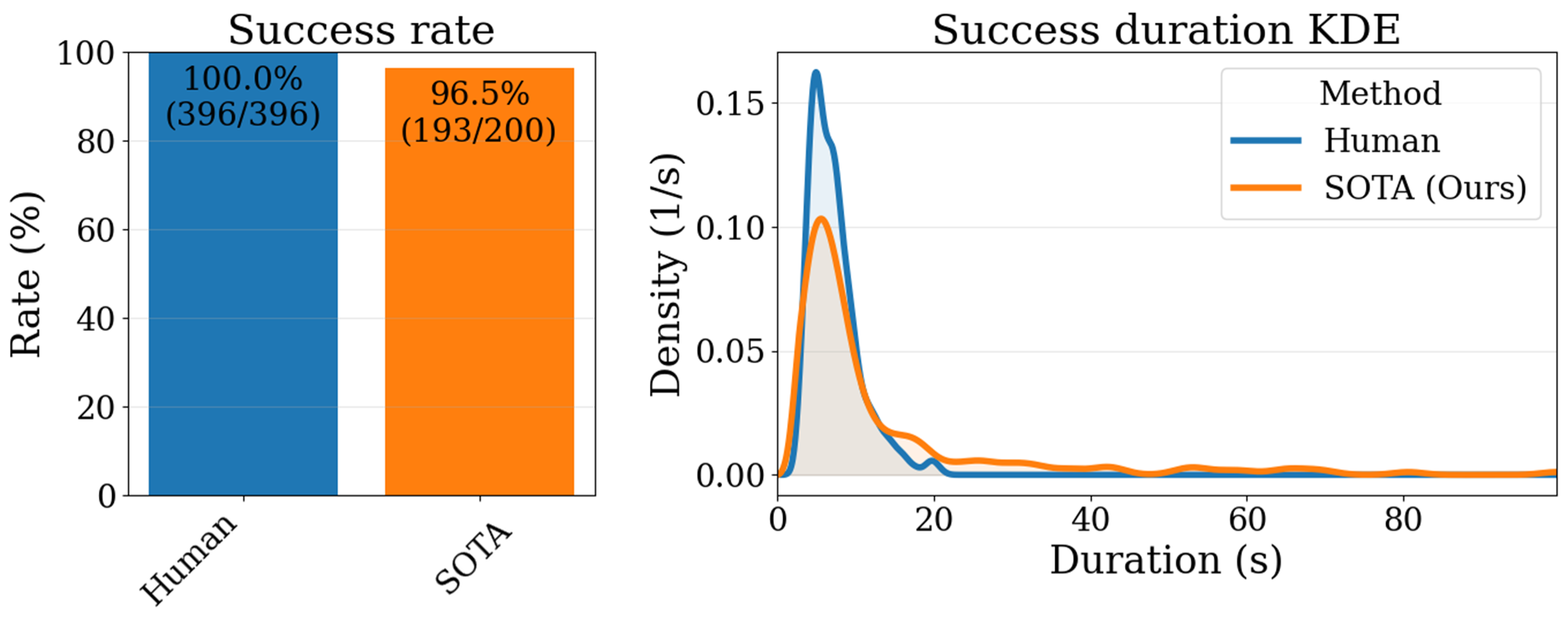}
\caption{BCM connector insertion: success rate (left) and success-only
completion-time density (right) of the SO-TA policy compared with human
demonstrations.}
\label{fig:bcm_speed}
\end{figure}

Across $200$ online rollouts the SO-TA policy attains a mean success-only
time close to human teleoperation ($\sim\!10~\mathrm{s}$), but exhibits
a long completion-time tail ($20$--$100~\mathrm{s}$) and $7$ timeouts
(Fig.~\ref{fig:bcm_speed}). This is the canonical compounding-error
pattern of imitation learning under contact: once the policy drifts off
the demonstration manifold, recovery is prolonged.

\paragraph*{Cutoff-and-reset wrapper}
\begin{figure}[!t]
\centering
\includegraphics[width=0.98\columnwidth]{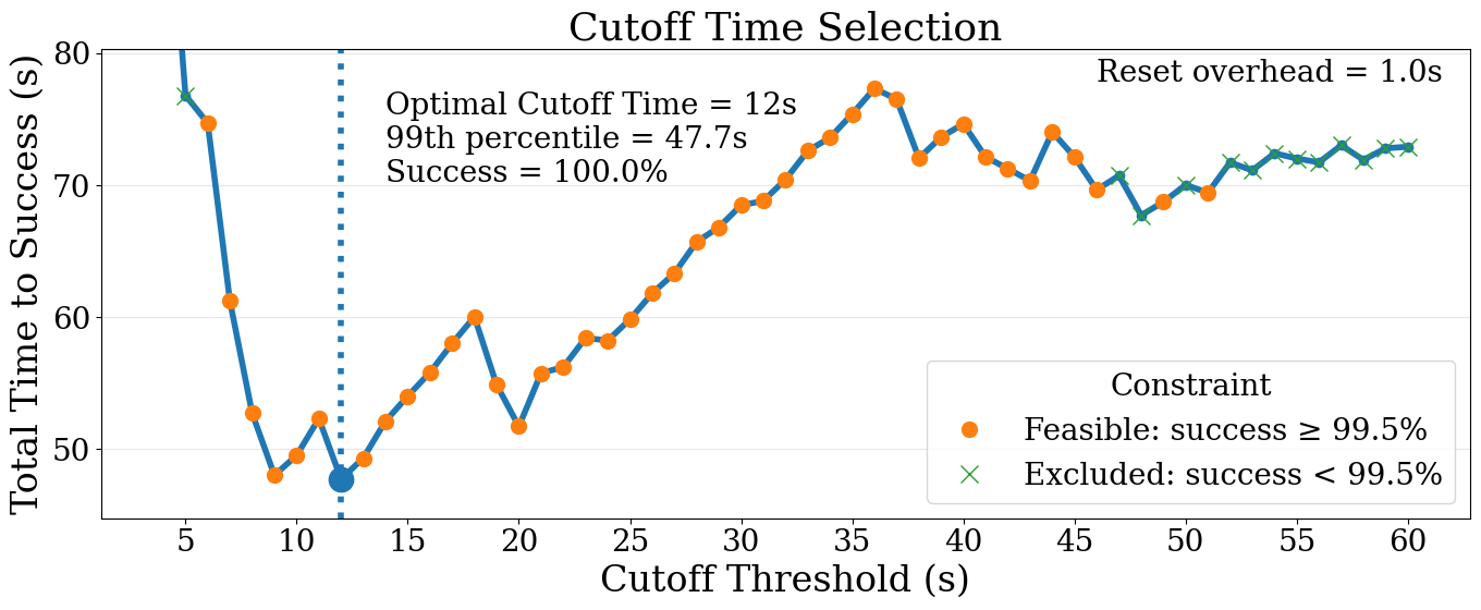}
\caption{Selecting the cutoff time by minimizing the $99^{\mathrm{th}}$
percentile of success-only total time under a $\ge99.5\%$ success
constraint. The selected cutoff is $12~\mathrm{s}$.}
\label{fig:bcm_cutoff_sel}
\end{figure}

\begin{figure}[!t]
\centering
\includegraphics[width=0.98\columnwidth]{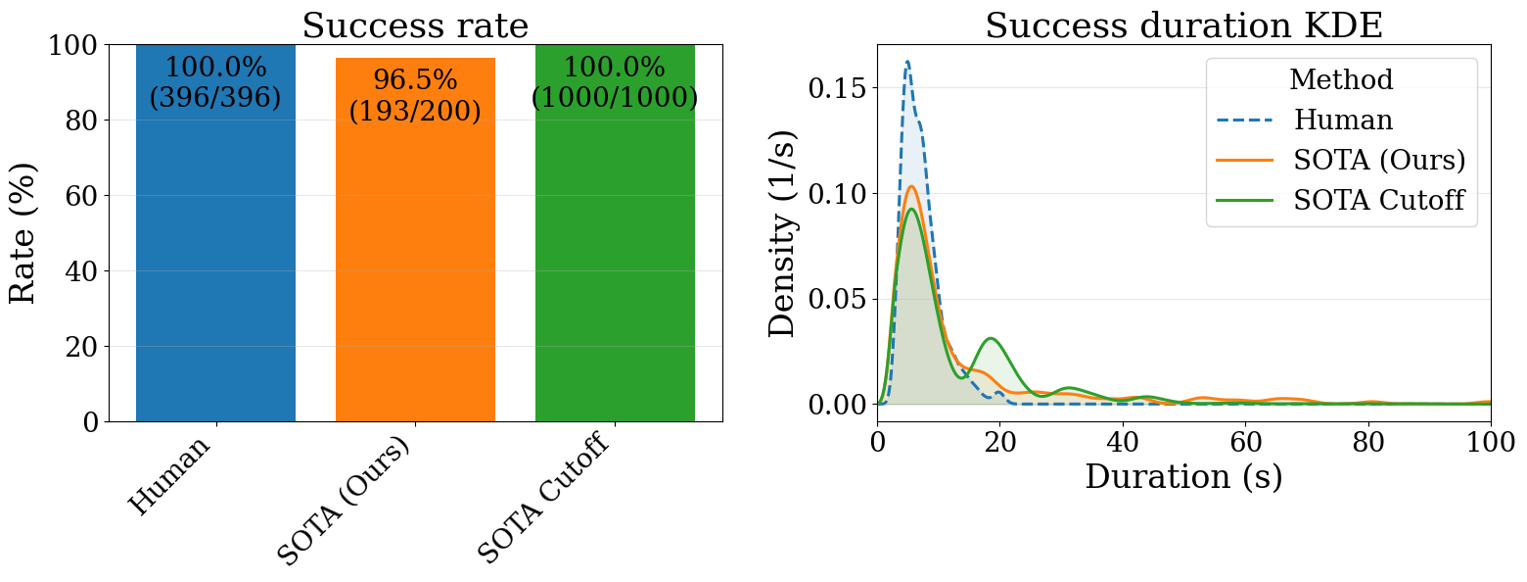}
\caption{With the $12~\mathrm{s}$ cutoff wrapper, the long-duration tail
is removed and the short-time mode is preserved.}
\label{fig:bcm_cutoff}
\end{figure}

We introduce a lightweight runtime wrapper that aborts and restarts a
rollout once it has not succeeded by time $t$, with a $1~\mathrm{s}$
reset overhead. Bootstrapping single-attempt outcomes under the
$100~\mathrm{s}$ budget, we select $t$ to minimize the
$99^{\mathrm{th}}$ percentile of total time-to-success while maintaining
$\ge\!99.5\%$ success (Fig.~\ref{fig:bcm_cutoff_sel}). The selected
cutoff $t\!=\!12~\mathrm{s}$ yields a $99^{\mathrm{th}}$-percentile total
time-to-success of $47.7~\mathrm{s}$ versus $70.85~\mathrm{s}$ without
cutoff, while reaching $100\%$ success in $1{,}000$ simulated episodes
(Fig.~\ref{fig:bcm_cutoff}). The wrapper improves tail latency without
modifying the trained policy.

\subsection{Curved-surface mark erasing}
\label{subsec:erase}

\begin{figure}[!t]
\centering
\includegraphics[width=0.95\columnwidth]{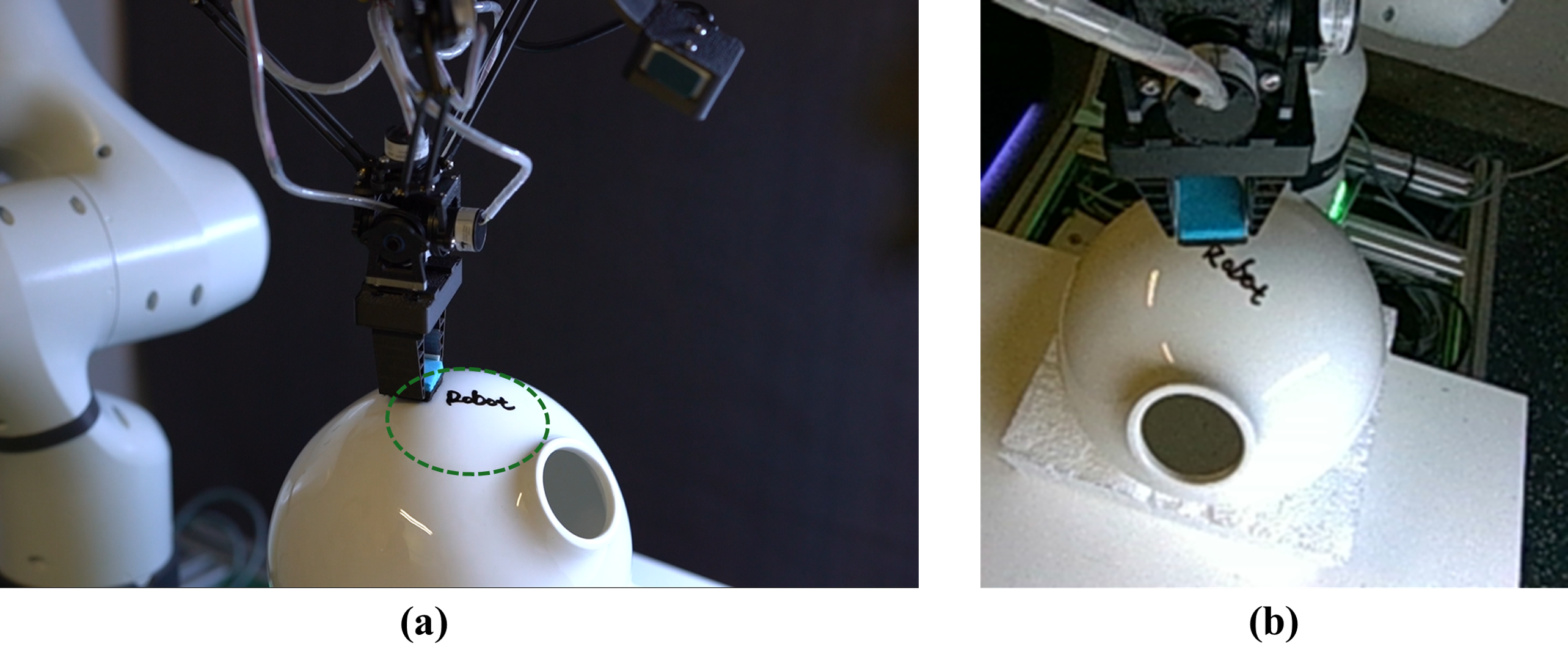}
\caption{Curved-surface mark erasing. (a) Setup with a handwritten
``Robot'' mark at a random location within a green region on the
vase. (b) Robot-camera view.}
\label{fig:erase_setup}
\end{figure}

Unlike insertion, mark erasing requires localizing the target on a
curved surface, establishing contact, regulating normal force, and
producing sustained sliding friction. The eraser frequently occludes the
remaining ink near the end of an episode, so short-horizon temporal
context matters. We collect $199$ demonstrations ($35{,}374$ steps,
$3{,}437.4~\mathrm{s}$, $31{,}991$ training pairs).

\begin{figure}[!t]
\centering
\includegraphics[width=0.98\columnwidth]{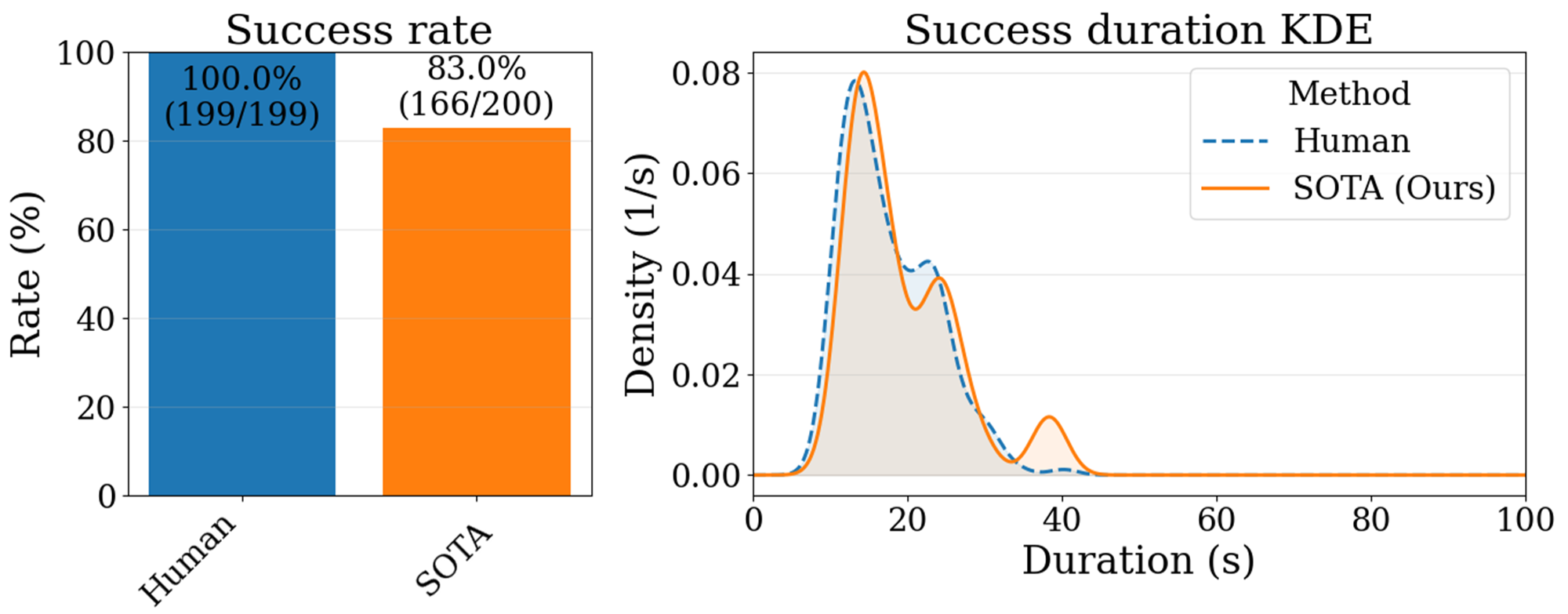}
\caption{Erasing: success rate (left) and success-only completion-time
density (right) of the SO-TA policy compared with human demonstrations.}
\label{fig:erase_result}
\end{figure}

Across $200$ online rollouts (Fig.~\ref{fig:erase_result}), SO-TA reaches
a mean success-only time of $19.3~\mathrm{s}$ versus $17.6~\mathrm{s}$
for human teleoperation, with $166/200$ successes under the strict
``all marks removed'' criterion. The KDE displays a multi-modal
structure consistent with $1$-, $2$- and $3$-attempt successes. Failures
correlate with mark location: under-represented boundary placements are
substantially more likely to require multiple attempts.

\begin{figure}[!t]
\centering
\includegraphics[width=0.98\columnwidth]{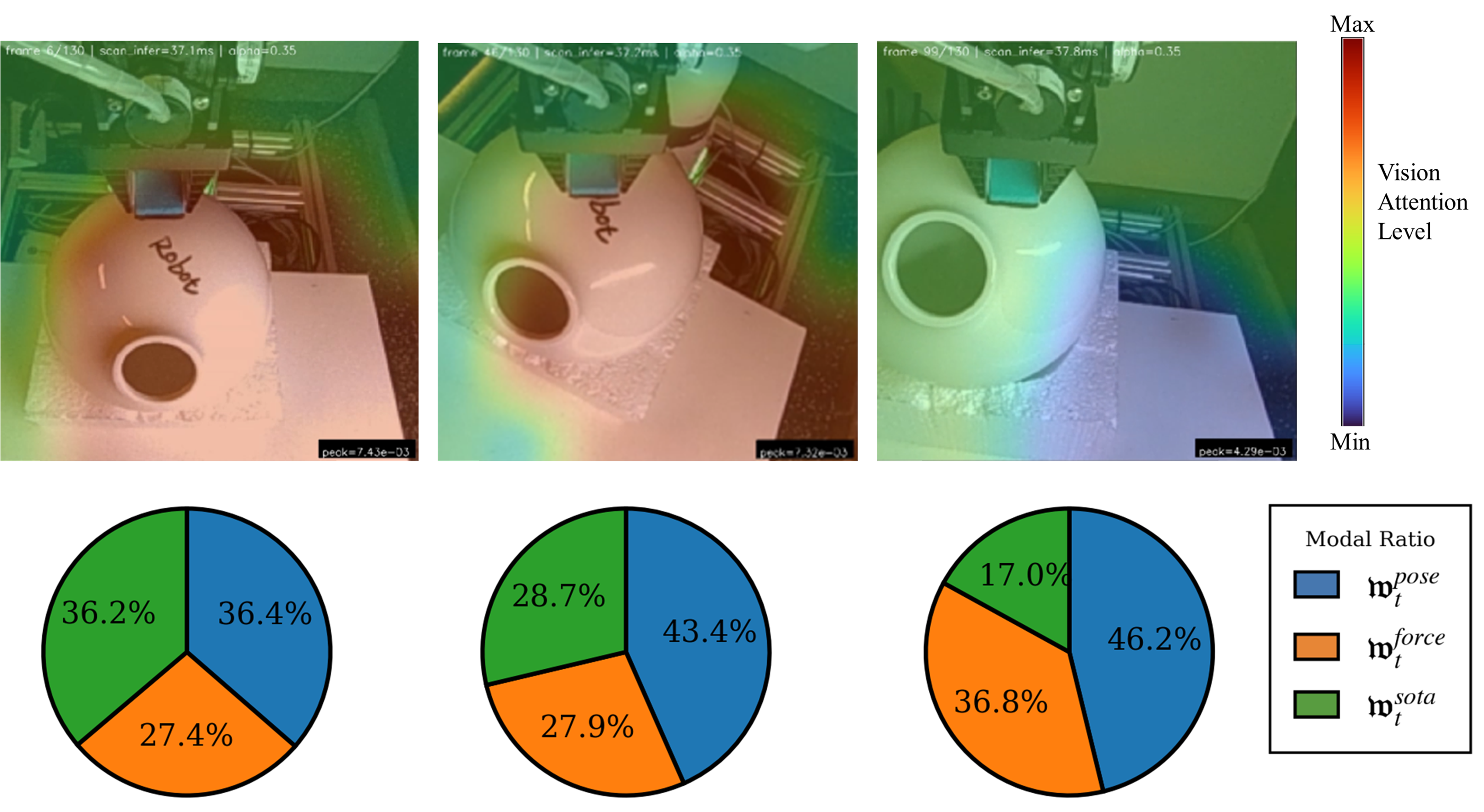}
\caption{Erasing: SO-TA interpretability across three phases. Vision
peaks before contact; visual and force shares are jointly emphasized
during wiping; pose dominates during retreat.}
\label{fig:erase_explain}
\end{figure}

The interpretability protocol of Sec.~\ref{subsec:peg} extends naturally
(Fig.~\ref{fig:erase_explain}). Pose remains consistently large
throughout the episode---likely a signature of the stereotyped operator
routine in the dataset. Before contact, vision attains its highest
share and force is near zero. During erasing the visual share decreases
but remains comparable to force, consistent with regulating contact
while choosing wiping direction. During retreat the visual share drops
sharply ($\sim\!17\%$) and pose dominates ($\sim\!46\%$), matching a
near-deterministic return motion. OT heatmaps concentrate on the vase
surface during approach, persist during erasing, and become more diffuse
near episode termination.

\section{Conclusion}
\label{sec:conclusion}
We presented \textbf{SO-TA}, a tri-modal fusion module that casts
force--pose-conditioned spatial aggregation as an entropy-regularized
Optimal Transport problem with explicit marginal constraints. Paired
with a diffusion-policy backbone, SO-TA matches or exceeds concatenation
and cross-attention baselines on three real-robot contact-rich tasks
under matched capacity, and retains $82.5\%$ success under visual
perturbation where concatenation drops to $43.5\%$. Two limitations
remain: OT heatmaps stay somewhat diffuse because the cost
in~\eqref{eq:cost} carries no spatial prior, and structured fusion shrinks
but does not eliminate the compounding-error long tail under contact,
which the cutoff-and-reset wrapper closes only at runtime. The broader
takeaway is that structured inductive biases on the fusion step remain
a high-leverage axis for visuo-haptic policy learning, even as
action-producing backbones scale.

\bibliographystyle{IEEEtran}
\bibliography{references}

%
%
%

\vfill

\end{document}